 \documentclass[pmlr,twocolumn,10pt]{jmlr} 

\usepackage{booktabs}
\usepackage{siunitx}

\usepackage{algpseudocode}
\usepackage{makecell}
\usepackage{multirow}
\usepackage{hyperref}
\usepackage{svg}
\usepackage{xcolor, colortbl}
\definecolor{Gray}{gray}{0.85}
\definecolor{LightBlue}{rgb}{0.87, 0.93, 0.96} 
\definecolor{DarkOrange}{rgb}{0.75,0.41,0} 
\definecolor{LightOrange}{rgb}{1,0.78,0.5} 

\usepackage[switch]{lineno}

\theorembodyfont{\upshape}
\theoremheaderfont{\scshape}
\theorempostheader{:}
\theoremsep{\newline}
\newtheorem*{note}{Note}

\jmlrvolume{LEAVE UNSET}
\jmlryear{2024}
\jmlrsubmitted{LEAVE UNSET}
\jmlrpublished{LEAVE UNSET}
\jmlrworkshop{Machine Learning for Health (ML4H) 2024} 

 \title[Leveraging Foundation Language Models for Automated Cohort Extraction from Large EHR Databases]{Leveraging Foundation Language Models (FLMs) for Automated Cohort Extraction from Large EHR Databases}

\author{%
\Name{Purity Mugambi} \Email{pmugambi@cs.umass.edu}\\
\addr University of Massachusetts, Amherst, USA
\AND
\Name{Alexandra Meliou} \Email{ameliou@cs.umass.edu}\\
\addr University of Massachusetts, Amherst, USA
\AND
\Name{Madalina Fiterau} \Email{mfiterau@cs.umass.edu}\\
\addr University of Massachusetts, Amherst, USA
}

\begin{document}

\maketitle

\begin{abstract}
A crucial step in cohort studies is to extract the required cohort from one or more study datasets. This step is time-consuming, especially when a researcher is presented with a dataset that they have not previously worked with. When the cohort has to be extracted from multiple datasets, cohort extraction can be extremely laborious. In this study, we present an approach for partially automating cohort extraction from multiple electronic health record (EHR) databases. We formulate the guided multi-dataset cohort extraction problem in which selection criteria are first converted into queries, translating them from natural language text to language that maps to database entities. Then, using FLMs, columns of interest identified from the queries are automatically matched between the study databases. Finally, the generated queries are run across all databases to extract the study cohort. We propose and evaluate an algorithm for automating column matching on two large, popular and publicly-accessible EHR databases -- MIMIC-III and eICU. Our approach achieves a high top-three accuracy of $92\%$, correctly matching $12$ out of the $13$ columns of interest, when using a small, pre-trained general purpose language model. Furthermore, this accuracy is maintained even as the search space (i.e., size of the database) increases.
\end{abstract}
\begin{keywords}
Cohort Extraction, Schema Matching, Language Models, EHR
\end{keywords}

\paragraph*{Data and Code Availability}
We evaluate the presented method on two large and publicly-accessible (after accepting a data usage agreement) electronic health records (EHR) datasets: MIMIC-III \cite{Johnson2016} and eICU \cite{Pollard2018}. 

\paragraph*{Institutional Review Board (IRB)}
This study does not require IRB approval.

\section{Introduction}
\label{sec:intro}
Cohort extraction is a crucial step in clinical and health informatics research, and especially, in observational retrospective exploratory studies. Typically, when conducting these studies researchers have access to one or more  databases from which they extract the cohort of interest using predetermined inclusion/exclusion criteria. It is also common for these studies to have a primary dataset from which the key findings are learned and a secondary dataset which is used to validate those findings. EHR databases are the main data sources for this research. 

EHR databases contain real-world data and are attractive for retrospective research because their data is abundant, immediately available, and researchers have high flexibility in terms of the kinds of analyses they can conduct on that data~\cite{Gehrmann2023}. However, data preparation costs are a huge challenge when using EHR data~\cite{Gehrmann2023}. Data heterogeneity~\cite{Sarwar2022, Kreuzthaler2015}, lack of standardization in clinical terms and database entities such as table and column names~\cite{Reisman2017-ea} and database sizes,  are some of the reasons data preparation is time consuming. For example, the popular EHR databases used in this study, MIMIC-III and eICU, have 26 and 31 tables, with a combined total of 324 and 391 columns respectively. In studies where researchers have to extract cohorts multiple times, such as, in health equity studies~\cite{Toseef2022, Claxton2019, Mugambi2023}, or in multi-dataset epidemiological studies~\cite{Knight2020}, data preparation costs can be extensive. 

We are motivated by the larger problem of health equity where researchers seek to compare outcomes of a specified cohort (e.g., patients with a diagnosis of kidney disease who received dialysis) across multiple datasets. Researchers would need to extract that cohort from each of the study datasets. We term this \emph{the \textbf{multi-dataset cohort extraction} problem} and define it as \emph{selecting patients that match some inclusion/exclusion criteria from multiple datasets that might have a different structure}. Multi-dataset cohort extraction is a laborious activity especially when a researcher is presented with a new database that they have no prior experience with.

This problem is at the intersection of cohort extraction (i.e., given a set of criteria, retrieve a specific cohort from the data) and schema matching (i.e., presented with a schema, e.g., column names and their values, find its matches in different databases). Several recent studies have presented approaches for automating patient cohort extraction/discovery~\cite{Yuan2019, Franklin2020-dg, Gokhale2020, Grardin2022}, and automating schema matching~\cite{Dong2021, Khatiwada2022, Suhara2022, Fan2023} indicating that this is still an important and unsolved problem.

Guided by prior work that has shown that language foundation models can greatly benefit data integration tasks~\cite{Narayan2022}, we sought to examine how pre-trained language models (LMs) can be leveraged to automate the extraction of specified patient cohorts from multiple databases. We formulated the multi-dataset cohort extraction problem as a three-step process; first, the criteria is converted into a set of easy-to-execute queries, second, pre-trained LMs are used to match columns of the databases used in the study, and third, the generated queries are executed on each of the databases. In this study, we developed and evaluated an algorithm for automating column matching -- the second part of the problem. Prior solutions can be used to convert criteria to queries (part one of the problem) as discussed in Section \ref{ssec:criteria-to-queries}.

This study makes two main contributions. First, we formulate the \textbf{guided} multi-dataset cohort extraction problem which bridges the gap between inclusion criteria and retrieval of a patient cohort from multiple datasets. Second, we propose, implement, and evaluate a column-matching algorithm using a small pre-trained LM on two large, popular, and publicly-accessible EHR databases. It is particularly suited to ML4H audience because it presents evidence of how deep language models can be utilized to quickly extract cohorts from multiple datasets -- a task that to the best of our knowledge is largely manual requiring many hours from clinical and ML experts. Even partial automation of cohort extraction can cut down on the data preparation time that significantly hinders multi-dataset analyses and create  opportunities for researchers to replicate their analyses on cohorts extracted from other datasets which, as argued by~\cite{Bakken2021}, is essential in increasing the impact of biomedical and health informatics findings.

\section{Related work}
\label{sec: related-work}
\textbf{Cohort discovery via disease phenotyping}: In this research area, patient cohorts are discovered in data through disease phenotyping. Cohorts can be manually curated from a specific criteria that is developed by advanced researchers for identifying a disease. They are often presented as a set of rules~\cite{Liao2015}. The criteria can also be converted into algorithms, aggregated and offered as the gold standard; for instance, PheKB~\cite{Kirby2016}. This approach is time consuming and has motivated the development of automated methods for cohort discovery which are largely based on representation learning. To discover cohorts, these models learn embeddings for each patient and then group patients with similar embeddings together (based on some distance metric) and/or use key medical concepts to retrieve patient groups. Due to the versatility of representation learning methods, automated tools for cohort discovery have been developed for a wide range of applications including; general disease risk prediction~\cite{Miotto2016}, identification of patient cohorts by learning disease phenotypes~\cite{Kandula2011, Yu2015, Glicksberg2018-vf}, identifying computational structures of multiple diseases~\cite{Pivovarov2015}, and learning cohort definitions as a side effect of bulk database phenotyping~\cite{Chiu2017}. 

The present study differs from this class of work in three ways. First, the problems, while related, are different. Research on cohort discovery learns a criteria that identifies a specific disease phenotype. We developed a method for retrieving a cohort from a database when presented with a selection criteria. Second, cohort discovery research learns patient embeddings and computes distances between those embeddings to identify similar patients. Our method computes embeddings of database columns to identify their matches in other databases. Third, most work in cohort discovery trains and optimizes a model for each dataset. We utilize a small FLM to learn embeddings across multiple datasets simultaneously.

A separate body of work that is closely related to cohort discovery is often referred to as ``cohort selection.'' Similar to our task, this research retrieves a cohort of interest from a clinical database. However, many papers published on this topic, all of which were submissions to the 2018 National Natural Language Processing (NLP) Clinical Challenges (n2c2) Shared-Task and Workshop on Cohort Selection for Clinical Trials, formulate cohort selection as a multi-labeling problem with the goal of identifying which criterion is met for each patient record~\cite{SeguraBedmar2019, Chen2019, Xiong2019, ChenLongGu2019}. They differ from the present study both in the problem formulation and methodology.

\textbf{Column search and matching}: A large field of research on data management has developed methods for searching through large data sources; often referred to as \emph{data lakes}~\cite{Wieder2022-kz, Hai2023}. Whereas some of this work relies on database operations such as join, union, and search \cite{Khatiwada2022, Zhang2020, Zhu2019}, much recent work, similar to this study, is on learning high dimensional representations of various components of the lake. Within this space sits research on using LMs for schema matching, often termed ``discovery of joinable tables". A large portion of this work~\cite{Li2020, Suhara2022, Dong2023} learns representations of the components of the lake by training/finetuning a large language model (LLM) in a supervised task. They leverage component-relatedness metrics such as table-joinability to apriori label data elements as positive or negative examples that can be utilized in supervised learning. A much smaller number of studies, more closely-related to this study, use LLMs in an unsupervised approach~\cite{Fan2023, Dong2021}. While methodologically similar, our study differs from this unsupervised work in application. Models from these papers were finetuned for and evaluated on large open data benchmarks such as OPEN~\cite{Zhu2016}, a dataset of relational tables from the Canadian open data repository, and the Web Data Commons (WDC) web table corpus~\cite{WDC2015}, that are structurally and contextually different from healthcare data.

\section{Methods}
\label{sec:methods}
\subsection{Problem decomposition}
\label{sec:problem}
This section discusses the problem components withholding implementation details which are discussed in Section~\ref{ssec:method}.

\textbf{Context}: A researcher has inclusion/exclusion criteria and multiple databases. The researcher has, at a minimum, a high-level understanding of the structure (i.e., tables, their columns, and data types) of at least one of the databases, henceforth referred to as the reference database. This is a reasonable assumption because the structure of the database is provided as metadata, and/or through the dataset documentation. Furthermore, many researchers use their institutional databases for observational studies and would typically know their structure. Additionally, technical support, including cohort discovery tools, are made available to assist researchers in quickly understanding the structure of many databases. The researcher wants to use the criteria to extract a cohort from the reference database(s) and other databases they have access to, henceforth referred to as unknown databases.

We formulate guided cohort extraction as a three-step problem: (1)~break down the criteria into a set of queries, (2)~match columns of interest in the~\textbf{reference} database(s) to their counterparts in the~\textbf{unknown} database(s), and (3)~execute the selection criteria queries on all databases to obtain the cohorts.

\subsubsection{Break down criteria to queries}
\label{ssec:criteria-to-queries}
Suppose the inclusion criteria is: \emph{patients with a diagnosis of Asthma, aged 20--35, who were prescribed inhaler corticosteroids}. A researcher could write three SQL queries; 
\begin{verbatim}
    1) SELECT * FROM <diagnosis-table>
     WHERE LOWER(<diagnosis-column>) = 
     `asthma';
    2) SELECT * FROM <patients-table> 
    WHERE <age-column> BETWEEN 20 AND 35;
    3) SELECT * FROM <prescriptions-table> 
    WHERE <drug-column> IN 
    <list of inhaled-corticosteroids>;
    \end{verbatim}
and merge the results through an inner-join command to obtain the required cohort. They could also write a single SQL query or even use no-SQL tools such as dataframes. The goal of this step is to translate the criteria into a language that~\textbf{contains} database structure and language. Translating criteria into queries is typical in observational research. As such, in this study we expect that the researcher has (or will build) the necessary skills to solve this part of the problem and do not develop any methods for it. However, researchers could also leverage tools that have been developed to automatically convert selection criteria into OMOP common data model~\cite{Hripcsak2015-gf} objects~\cite{Kang2017} or SQL queries~\cite{Yuan2019, yu-etal-2020-dataset}.  

\subsubsection{Match columns of interest}
From step one above, we have translated the criteria into tables and columns. In this step, we first identify those tables and columns in the reference database, then, we find the best match(es) for each of the identified columns from the unknown database(s). For instance, using the first query from step one above, the researcher would look through their reference database and find which table contains diagnosis information, and which column in that table has the diagnosis text. If the diagnosis information is in a column named \emph{diagnosisText}, in this step, the best match (e.g., \emph{diagnosis}) would be obtained from an unknown database. This study develops and evaluates an algorithm to match columns across databases, which is discussed in detail in Section~\ref{ssec:method}.

\subsubsection{Execute queries on all databases}
Finally, since all the relevant tables and columns of the \emph{unknown} database have been identified, the queries developed in part one can be executed to obtain the cohort. This problem is also typical in observational studies, therefore we do not develop methods for it.

Formulating the problem in this way allows us to:
\begin{enumerate}
    \item leverage the researcher's knowledge of their current database to rapidly understand the new unknown database(s), lowering the barrier for multi-dataset analyses.
    \item use the structure of the known database~\textbf{as a guide} to find relevant tables and columns in unknown databases. This reduces errors and computation time, which would otherwise occur if one was searching unguided across the entire database.
\end{enumerate}

In this paper, we propose a solution for the second part of the problem; i.e., automating column matching, and evaluate it in isolation. In the future, the proposed solution would be used as part of a complete system that also includes existing solutions for parts $1$ and $3$.

\subsection{Column matching}
\label{ssec:method}
\subsubsection{Algorithm}
We developed a column matching algorithm (Algorithm~\ref{algorithm-1}) that takes as input a list of all the columns from all the databases; reference and unknown. For each column, using a pre-trained LM, we generate a vector embedding of its values then aggregate them into a mean embedding for that column (line 2). Next, we compute similarity scores (i.e., distances between the mean vector embeddings) of all columns using cosine-similarity (equation~\ref{eqn1}, line 3) and return the best $k$ matches (line 4). $k$ is a user provided integer argument specifying how many top matches should be returned.

\begin{algorithm2e}[htbp]
\caption{Top-k column matching}
\label{algorithm-1}
\KwIn{List of columns $\mathcal{L}$}
\begin{algorithmic}[1]
\Require FLM $\mathcal{M}$
\Require CosineSim $\mathcal{CS}$
\end{algorithmic}
\begin{enumerate}
    \item $embed\_list = []$\;
    \item \For{col in $\mathcal{L}$}{
    \begin{enumerate}
        \item $values\gets col.values()$\;
        \item $col.emb\gets M(values)$\;
        \item $col.mean\_emb\gets SUM(col.emb)/len(values)$\;
        \item $embed\_list.append(col.mean\_emd)$\;
    \end{enumerate}
    }
    \item $distances\gets CS(embed\_list)$\;
    \item $best\_match\_by\_values \gets top\-K(distances, k)$\;
    \item $m\_embed\_list = []$\;
    \item $top\_matches \gets top\_K(distances, threshold)$\;
    \item \For{match in $top\_matches$}{
    \begin{enumerate}
        \item $match.metadata.emb\gets M(match.metadata)$\;
        \item \parbox[t]{4cm}{$m\_embed\_list.append\\(match.metadata.emb)$}\;
    \end{enumerate}
    }
    \item $m\_distances \gets CS(metadata\_embed\_list)$\;
    \item $best\_matches\_by\_m \gets top\_K(m\_distances, k)$\;
\end{enumerate}
\end{algorithm2e}

\begin{equation}\label{eqn1}
    cossim(x,y) = \frac{x^Ty}{||x||||y||}
\end{equation}

Whereas column data is most informative when finding matches, metadata (such as column name and data type) could provide extra context essential in differentiating between close matches. To examine the effect of metadata in disambiguating between matches, we first obtain top $M$ matches whose similarity distance is greater than or equal to some threshold (line 6). Then, for each of these matches, we separately encode the column name, its datatype, and table names for which it appears (line 7). Finally, we compute similarity scores based on the metadata embedding vectors and return the best $k$ matches (lines 8-9). The computational efficiency of the algorithm is linear in the number of columns, $\mathcal{L}$.

\subsubsection{Implementation}
Code was written using Python 3.10. We used the Bi-directional Encoder Representations from Transformers (BERT) FLM to generate the embeddings. BERT can be accessed in various forms; we chose sentence-BERT (SBERT)~\cite{reimers-2019-sentence-bert} because it contains the state of the art image and text embeddings and is readily available as a module in Python. Within SBERT, we picked the pre-trained \emph{all-MiniLM-L6-v2} model because it was trained on a large dataset (1 billion training pairs) and is considered a general purpose use model~\cite{sbert-documentation}. Unlike other general purpose models trained on the same size of data,~\emph{all-MiniLM-L6-v2}, henceforth referred to as SBERT-model, is five times faster~\cite{sbert-documentation}. 

The SBERT-model generates a $384$-dimensional vector embedding for each input. In this study, we treated each column value as an independent sentence and obtained a vector embedding for it. To reduce memory usage and computation time we only encoded unique values in each column. Whereas SBERT is trained to encode text/image data, we hypothesized that it would generate semantically meaningful embeddings for numerical data, and used it to also encode floats, dates, and integers. Prioritizing the development of a flexible method that could deal with data variety, we chose to use one model (i.e., SBERT-model) and cast everything to text, especially because our prior experience with MIMIC had showed us that several columns can have mixed numerical and text data.

We ran the models on a compute cluster with $100-500$GB RAM and $2-5$ GPUs. Despite only encoding unique values in each column, we processed input data in chunks to ensure that the raw data and their corresponding large embeddings could fit in memory. Data was processed using the Pandas library and intermittent results written to HDF5 files.

\subsection{Experimental setup}
\label{ssec-setup}
We conducted experiments on two large, publicly-accessible EHR datasets; MIMIC-III and eICU, with MIMIC-III as the reference database and eICU the unknown. To evaluate the accuracy of the match algorithm, we manually identified matches for columns in MIMIC-III in eICU. As stated in Section~\ref{sec:problem}, this method would be applied to select a cohort based on a criteria, typically to answer a specific research question. Therefore, to evaluate the proposed solution, we developed the usecase below. 

\textbf{Research question}: Are there differences in pharmacological treatment of acute myocardial infarction (AMI) by patient sex, race, age, and type of insurance?

\textbf{Inclusion criteria}: Adult patients with a primary diagnosis of AMI who have no known comorbidities and were prescribed drugs in at least one of the four drug-classes for standard pharmacological treatment of AMI; Angiotensin-converting-enzyme (ACE) inhibitors, statins, beta-blockers and anticoagulants/antiplatelets.

By breaking down the inclusion criteria into queries, we identified 
\textit{\textless diagnosis-table\textgreater},~\textit{\textless diagnosis-column\textgreater},~\textit{\textless diagnosis-priority-column\textgreater},~\textit{\textless past-medical-history-table\textgreater},~\textit{\textless past-medical-history-value-column\textgreater},~\textit{\textless prescriptions-table\textgreater},~\textit{\textless drug-name-column\textgreater},~\textit{\textless patient-information-table\textgreater} and~\textit{\textless age-column\textgreater} as the tables and columns important to meet the inclusion criteria. In addition,~\textit{\textless sex-column\textgreater},~\textit{\textless insurance-column\textgreater} and~\textit{\textless race/ethnicity-column\textgreater} are also required to answer the question. The names of these tables and columns in MIMIC-III are provided in Table~\ref{table:1}, and their descriptions in Appendix Table~\ref{appendix-table:1}.

\begin{table}[htbp]
\caption{Names of identified table and columns of interest in MIMIC-III}
\centering
\resizebox{0.5\textwidth}{!}{\begin{tabular}{|c|c|}
\hline
\textbf{Symbol} & \textbf{Name in MIMIC-III} \\
\hline \hline
\textit{\textless diagnosis-table\textgreater} & diagnoses\_icd \\
\hline
\textit{\textless diagnosis-column\textgreater} & icd9\_code \\
\hline
\textit{\textless diagnosis-priority-column\textgreater} & seq\_num \\
\hline
\textit{\textless past-medical-history-table\textgreater}, & chartevents \\
\hline
\textit{\textless past-medical-history-value-column\textgreater} & value \\
\hline
\textit{\textless prescriptions-table\textgreater} & prescriptions \\
\hline
\textit{\textless drug-name-column\textgreater} & drug \\
\hline
\textit{\textless patient-information-table\textgreater} & patients \\
\hline
\textit{\textless age-column\textgreater} & (admittime - dob)$^{\mathrm{a}}$ \\
\hline
\textit{\textless sex-column\textgreater} & gender \\
\hline
\textit{\textless insurance-column\textgreater} & insurance \\
\hline
\textit{\textless race/ethnicity-column\textgreater} & ethnicity \\
\hline
\multicolumn{2}{p{320pt}}{$^{\mathrm{a}}$ ``age" does not exist as a specific column in MIMIC-III. It is computed using the date of birth (dob) and hospital admission time (admittime).} \\
\end{tabular}}
\label{table:1}
\end{table}
 
\subsection{Experiments}
We first conducted experiments with a total of $52$ columns; columns of interest from MIMIC-III and their equivalent matches from eICU, ($32$ in total and shown in Table~\ref{table:2}), and a further $20$ randomly-selected columns from eICU. We then increased the number of randomly-selected columns from eICU to $30$, $50$, $70$, $90$, $100$, and eventually all columns to examine how the results change when a few reference columns need to be matched against an increasing size of the unknown database. When including metadata in the matching, we set the threshold for top-ranked matches based on column-values alone (line 6 in Algorithm \ref{algorithm-1}) to the lowest value for which the column had matches. Experiments were repeated for $k \in [1-3]$.

\subsubsection*{Evaluation metric}
Similar to prior studies, such as \cite{Dong2023}, we evaluate the accuracy of the algorithm by determining whether the ``true" match is returned for each column. Whereas most prior work \cite{Dong2021, Fan2023, Dong2023} is evaluated on web datasets where a column can have multiple positive matches, our application expects a single match from the unknown database for each column in the reference database. This is because EHR databases are typically normalized and rely on foreign keys to connect tables unlike web datasets where data files are typically standalone. Therefore, our problem is harder which led us to compute accuracy at $k=1$, $k=2$ and $k=3$. That is, we assign a $score=1$ if the ``true" match is among the returned matches for each value of $k$. This way, $k=1$ is the hardest case, and the problem gets easier with increasing values of $k$. While $k=1$ is the ideal solution, small values of $k$ (for instance, $k \in [2-10]$) significantly reduce the search space for the researcher. We report the number of columns that were correctly matched for each value of $k$, e.g., $10/13$. 

\begin{table}[htbp]
\caption{Reference columns and their matches in eICU.}
\centering
\resizebox{0.5\textwidth}{!}{\begin{tabular}{|l|l|l|}
\hline
\textbf{\makecell{MIMIC\\column name}} & \textbf{\makecell{eICU\\column name}} &  \textbf{\makecell{eICU\\table name}}\\
\hline
\hline
subject\_id $^{\mathrm{a}}$ & uniquepid & patient\\
\hline
hadm\_id $^{\mathrm{a}}$ & patientunitstayid $^{\mathrm{a}}$ & patient\\
\hline
admittime & hospitaladmittime24 & patient\\
\hline
admission\_location & hospitaladmitsource & patient\\
\hline
discharge\_location & hospitaldischargelocation & patient\\
\hline
insurance $^{\mathrm{b}}$ &  - & - \\
\hline
ethnicity & ethnicity & patient \\
\hline
diagnosis & diagnosisstring & diagnosis\\
\hline
seq\_num & diagnosispriority $^{\mathrm{c}}$ & diagnosis\\
\hline
icd9\_code & icd9code & diagnosis\\
\hline
value & pasthistoryvaluetext & pasthistory \\
\hline
gender & gender & patient \\
\hline
dob $^{\mathrm{d}}$ &  - & - \\
\hline
drug & drugname & medication \\
\hline
route & routeadmin & medication \\
\hline
dose\_val\_rx & dosage & medication \\
\hline
\multicolumn{3}{p{320pt}}{$^{\mathrm{a}}$ These columns are patient identifiers and typically appear in several tables, but the data format is maintained across the tables.} \\
\multicolumn{3}{p{320pt}}{$^{\mathrm{b}}$ Insurance information  is not available in eICU, so there are no ``true" matches.} \\
\multicolumn{3}{p{320pt}}{$^{\mathrm{c}}$ Diagnosis priority is recorded through sequence numbers (integers) in MIMIC-III, however, this same information in eICU is recorded as short text (string). They contain the necessary information to identify diagnosis priority but are very different structurally.} \\
\multicolumn{3}{p{320pt}}{$^{\mathrm{d}}$ Age is not pre-computed and stored as a column in MIMIC-III. It is instead computed by subtracting two timestamps (admittime and dob). eICU does not have date of birth (dob) data, instead it has an `age' column in the patient table.} \\
\end{tabular}}
\label{table:2}
\end{table}

\section{Results}
\label{sec: results}
\subsection{Summary of results}

\begin{figure}[htbp]
\includegraphics[width=\columnwidth]{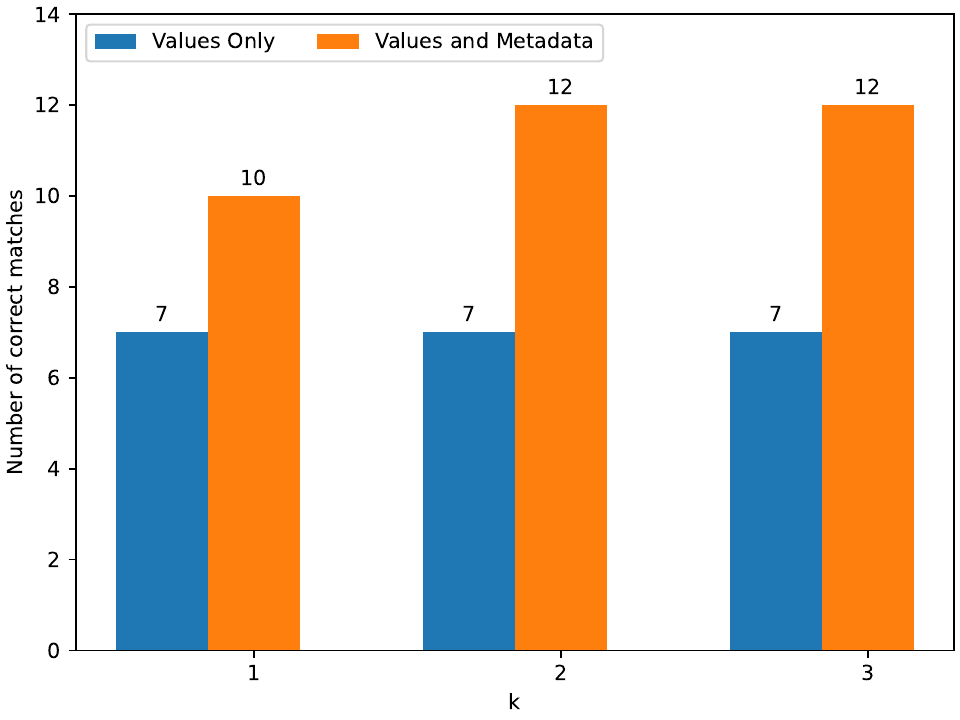}
\caption{Match accuracy for various values of $k$. $N$(total number of columns evaluated) $=13$}
\label{fig1}
\end{figure}

When running experiments with $52$ columns (i.e., only $20$ additional columns are selected from eICU), the SBERT-model correctly identified $7$ out of the $13$ columns ($53.8\%$; blue bars in Fig.\ref{fig1}) among the top-$3$ matches when matching using column values only, and $12$ ($92.3\%$; orange bars in Fig.\ref{fig1}) when metadata is used to sort among matches that have a similarity score of $0.4$ and above. Top-$3$ matches when using column values only are shown in Table~\ref{table:3} and when using values and metadata in Table~\ref{table:4}. While a total of $16$ columns were identified using the developed usecase, accuracy was computed only on the $13$ columns that have an exact match in eICU. The three that do not have matches are shaded in gray in the result Tables. In Table~\ref{table:4}, the metadata used to obtain the similarity score for each match is indicated. 

\begin{table}[htbp]
\caption{Match results using column mean vector embedding}
\centering
\resizebox{0.5\textwidth}{!}{\begin{tabular}{|l|l|l|l|}
\hline
\textbf{Ref-col} & \textbf{Match1} & \textbf{Match2} & \textbf{Match3} \\
\hline
\rowcolor{LightBlue}
admittime & admittime24$^{\mathrm{a}}$ & uniquepid & intakeoutputid\\
\hline
\rowcolor{LightBlue}
\makecell{admission\\location} &  admitsource$^{\mathrm{a}}$ & allergynotetype & admitdiagnosis\\
\hline
\rowcolor{LightBlue}
\makecell{discharge\\location} & dischargelocation$^{\mathrm{a}}$ & admitsource$^{\mathrm{a}}$  & admitdiagnosis\\
\hline
\rowcolor{LightBlue}
ethnicity &  ethnicity & admitdiagnosis & routeadmin \\
\hline
\rowcolor{LightBlue}
diagnosis &  diagnosisstring & admitdxtext & pasthistoryvalue \\
\hline
\rowcolor{LightBlue}
drug & drugname & admitdiagnosis & dosage \\
\hline
\rowcolor{LightBlue}
route & routeadmin & admitdiagnosis & frequency \\
\hline
\rowcolor{Gray}
seq\_num &  meanbp & age & fio2\\
\hline
\rowcolor{Gray}
dob & uniquepid & admittime24$^{\mathrm{a}}$ & customlabid\\
\hline
\rowcolor{Gray}
insurance & admitsource$^{\mathrm{a}}$ & dischargelocation$^{\mathrm{a}}$ & admitdiagnosis\\
\hline
\rowcolor{LightOrange}
subject\_id & customlabid & intakeoutputid & patientunitstayid \\
\hline
\rowcolor{LightOrange}
hadm\_id & intakeoutputid & customlabid & nursingchartid \\
\hline
\rowcolor{LightOrange}
icd9\_code & customlabid & patientunitstayid & noteid \\
\hline
\rowcolor{LightOrange}
value & labothervaluetext & gender & frequency \\
\hline
\rowcolor{LightOrange}
gender & teachingstatus & routeadmin & admitdiagnosis \\
\hline
\rowcolor{LightOrange}
dose\_val\_rx & gender & labothervaluetext & diabetes \\
\hline
\multicolumn{4}{p{320pt}}{$^{\mathrm{a}}$ These columns names have ``hospital" preceding them in the database.} \\
\multicolumn{4}{p{330pt}}{Shading; light blue: correct, light orange: incorrect, gray: no exact matches.} \\
\end{tabular}}
\label{table:3}
\end{table}

\begin{table}[htbp]
\caption{Match results with mean embedding enriched with metadata}
\centering
\resizebox{0.5\textwidth}{!}{\begin{tabular}{|l|l|l|l|}
\hline
\textbf{Ref-col} & \textbf{Match1} & \textbf{Match2} & \textbf{Match3} \\
\hline
\rowcolor{LightBlue}
hadm\_id & patientunitstayid & noteid & intakeoutputid \\
\rowcolor{LightBlue}
& CN, DT, TN & CN, DT & CN, DT \\
\hline
\rowcolor{LightBlue}
admittime & admittime24$^{\mathrm{a}}$ & uniquepid & icd9code\\
\rowcolor{LightBlue}
& CN, DT & CN, DT, TN & CN, DT \\
\hline
\rowcolor{LightBlue}
\makecell{admission\\location} &  admitsource$^{\mathrm{a}}$ & allergynotetype & -\\
\rowcolor{LightBlue}
& CN, DT & CN, DT & - \\
\hline
\rowcolor{LightBlue}
\makecell{discharge\\location} & drugname & dischargelocation$^{\mathrm{a}}$  & admitdxtext\\
\rowcolor{LightBlue}
& CN, DT, TN & CN, DT, TN & CN, DT, TN \\
\hline
\rowcolor{LightBlue}
ethnicity &  ethnicity & drugname & frequency \\
\rowcolor{LightBlue}
& CN & CN, DT, TN & CN, DT, TN \\
\hline
\rowcolor{LightBlue}
diagnosis &  diagnosisstring & admitdiagnosis & drugname \\
\rowcolor{LightBlue}
& CN, DT & CN, DT & CN, DT, TN \\
\rowcolor{LightBlue}
\hline
icd9\_code & icd9code & noteid & patientunitstayid \\
\rowcolor{LightBlue}
& CN & CN, DT & CN, DT \\
\hline
\rowcolor{LightBlue}
gender & gender & ethnicity & teachingstatus \\
\rowcolor{LightBlue}
& CN & CN, DT, TN & CN, DT, TN \\
\hline
\rowcolor{LightBlue}
drug & drugname & dosage & frequency \\
\rowcolor{LightBlue}
& CN, DT & CN, DT, TN & CN, DT, TN \\
\hline
\rowcolor{LightBlue}
route & routeadmin & dosage & frequency \\
\rowcolor{LightBlue}
& CN, DT, TN & CN, DT, TN & CN, DT, TN \\
\hline
\rowcolor{LightBlue}
dose\_val\_rx & dosage & frequency & routeadmin \\
\rowcolor{LightBlue}
& CN, DT, TN & CN, DT, TN & CN, DT, TN \\
\hline
\rowcolor{LightBlue}
value & labothervaluetext & pasthistoryvalue & frequency \\
\rowcolor{LightBlue}
& CN, DT & CN, DT & CN, DT \\
\hline
\rowcolor{Gray}
insurance & drugname & admitdxtext & admitsource$^{\mathrm{a}}$ \\
\rowcolor{Gray}
& CN, DT, TN & CN, DT, TN & CN, DT, TN \\
\hline
\rowcolor{Gray}
seq\_num &  age & fio2 & icd9code\\
\rowcolor{Gray}
& CN, DT & CN, DT & CN, TN \\
\hline
\rowcolor{Gray}
dob & admittime24$^{\mathrm{a}}$ & uniquepid & icd9code\\
\rowcolor{Gray}
& CN, DT, TN & CN, DT, TN & CN, DT, TN \\
\hline
\rowcolor{LightOrange}
subject\_id & patientunitstayid & noteid & gender \\
\rowcolor{LightOrange}
& CN, DT, TN, & CN, DT & CN, DT \\
\hline
\multicolumn{4}{p{320pt}}{$^{\mathrm{a}}$ These columns names have ``hospital" preceding them in the database.} \\
\multicolumn{4}{p{330pt}}{Shading; light blue: correct, light orange: incorrect, gray: no exact matches. Metadata; CN: column name, TN: table name(s), DT: column data type } \\
\end{tabular}}
\label{table:4}
\end{table}

\begin{figure}[htbp]
\includegraphics[width=\columnwidth]{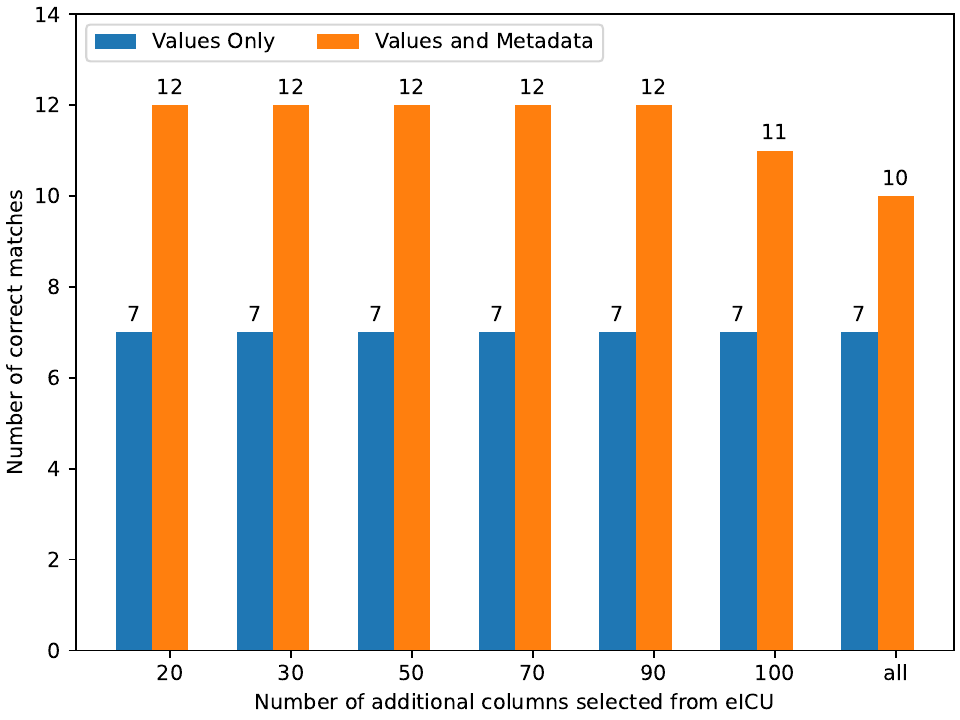}
\caption{Accuracy of top $3$ matches as the number of additional columns increases. $N$(total number of columns evaluated) $=13$}
\label{fig2}
\end{figure}

\subsection{Key takeaways}
\textbf{FLMs are suited for automated column matching.} Through a small general-purpose pre-trained LM, we achieve high accuracy in column matching. Despite being trained on non-medical, text and image data, SBERT-model shows good performance for healthcare data and for non-text data types, specifically, dates and integers. Surprisingly, the performance improves only slightly when $k>1$ indicating that SBERT-model is able to rank the ``true" match at the top.

\textbf{Metadata often improve match accuracy.} Inclusion of metadata can be useful in finding matches that are difficult to find using column values alone. Examples of columns that benefited from inclusion of metadata, \emph{icd9\_code}, \emph{gender}, and \emph{dose\_val\_rx}, have significantly different data formats between the two databases. In MIMIC-III, ICD9 codes are recorded as integers while in eICU they are a string combination of ICD 9\&10 codes. For instance, in MIMIC-III, ICD9 code~\emph{``410.71"} is recorded as~\emph{``41071"} while in eICU it is recorded as~\emph{``410.71, I21.4"}. Similarly, in MIMIC-III, gender is recorded as~\emph{`F'} or~\emph{`M'}, while in eICU it is~\emph{``female"} or~\emph{``male"}. Using the column name the algorithm is able to pick up the correct matches for both of these columns from a list of other close matches. Metadata do not always guarantee improvement in performance, as shown by~\emph{subject\_id}. When using values only, the LM is able to rank the correct match among the top three but when metadata is included it fails. The inclusion of extra information on column types and associated table names causes the algorithm to prioritize columns in eICU that appear in as many tables whose embeddings are close enough to the tables~\emph{subject\_id} appears in.

\textbf{Metadata protect against decreased performance as the database grows.} While the accuracy decreases as the size of the database grows, including metadata ensures that the accuracy is still relatively high, Fig.~\ref{fig2}.

\textbf{Column names alone are insufficient for column matching.} While it may appear as though column names alone could help us achieve high match accuracy, that is not the case. As shown in Table~\ref{table:4}, column names by themselves only helped match $3$ of the $13$ columns; ethnicity, icd\_code and gender because these are named in the exact same way in both databases. However, for the rest of the columns matching by column names alone produces a large number of false positives, as shown in Appendix Table~\ref{appendix-table:2}. Some of the reasons for this include: 1) short columns names (e.g., dob, drug) often have a high match rate with other columns because fewer characters are being used to match the sentences; 2) many columns can have a shared substring e.g., \emph{drug} and \emph{drugrate} have “drug” in them and would probably be matched by their column names despite containing different information; 3) columns can have similar names but represent different types of data. For instance, \emph{dischargelocation} and \emph{hospitaldischargelocation} are both columns in eICU representing the same information but in different data types. 

\subsection{Limitations and future work} This paper has not examined variance in algorithm efficiency because we believe it is out of scope and should be handled in a separate study. We also plan to examine how the algorithm performs when presented with a larger set of reference columns, for instance, by constructing and evaluating multiple selection criteria. Future work could also compare the performance of general-purpose LMs against LMs trained or fine-tuned on clinical data.

\section{Conclusion}
In this work, we presented an approach for automating matching of columns of EHR databases. Our results show that \textbf{the proposed approach is a good solution for automated column matching}. Not only is the match accuracy high, this performance persists even as the size of the unknown database increases. Additionally, our method can be \emph{run on clusters with limited compute resources} (2GPUs and 100GB RAM) and requires that \emph{a small FLM be run \textbf{just once}} to compute the embeddings. To obtain column matches, embeddings are looked up (from storage) and distances between them computed, which is an inexpensive operation. 

\bibliography{main}

\begin{thebibliography}{45}
\providecommand{\natexlab}[1]{#1}
\providecommand{\url}[1]{\texttt{#1}}
\expandafter\ifx\csname urlstyle\endcsname\relax
  \providecommand{\doi}[1]{doi: #1}\else
  \providecommand{\doi}{doi: \begingroup \urlstyle{rm}\Url}\fi

\bibitem[sbe()]{sbert-documentation}
URL \url{https://sbert.net/docs/sentence_transformer/pretrained_models.html#original-models}.

\bibitem[Bakken(2021)]{Bakken2021}
Suzanne Bakken.
\newblock Replication studies and diversity, equity, and inclusion strategies are critical to advance the impact of biomedical and health informatics.
\newblock \emph{Journal of the American Medical Informatics Association}, 28\penalty0 (9):\penalty0 1813–1814, August 2021.
\newblock ISSN 1527-974X.
\newblock \doi{10.1093/jamia/ocab168}.
\newblock URL \url{http://dx.doi.org/10.1093/jamia/ocab168}.

\bibitem[Chen et~al.(2019{\natexlab{a}})Chen, Warikoo, Chang, Chen, and Hsu]{Chen2019}
Chi-Jen Chen, Neha Warikoo, Yung-Chun Chang, Jin-Hua Chen, and Wen-Lian Hsu.
\newblock Medical knowledge infused convolutional neural networks for cohort selection in clinical trials.
\newblock \emph{Journal of the American Medical Informatics Association}, 26\penalty0 (11):\penalty0 1227–1236, August 2019{\natexlab{a}}.
\newblock ISSN 1527-974X.
\newblock \doi{10.1093/jamia/ocz128}.
\newblock URL \url{http://dx.doi.org/10.1093/jamia/ocz128}.

\bibitem[Chen et~al.(2019{\natexlab{b}})Chen, Gu, Ji, Lou, Sun, Li, Gao, and Huang]{ChenLongGu2019}
Long Chen, Yu~Gu, Xin Ji, Chao Lou, Zhiyong Sun, Haodan Li, Yuan Gao, and Yang Huang.
\newblock Clinical trial cohort selection based on multi-level rule-based natural language processing system.
\newblock \emph{Journal of the American Medical Informatics Association}, 26\penalty0 (11):\penalty0 1218–1226, July 2019{\natexlab{b}}.
\newblock ISSN 1527-974X.
\newblock \doi{10.1093/jamia/ocz109}.
\newblock URL \url{http://dx.doi.org/10.1093/jamia/ocz109}.

\bibitem[Chiu and Hripcsak(2017)]{Chiu2017}
Po-Hsiang Chiu and George Hripcsak.
\newblock Ehr-based phenotyping: Bulk learning and evaluation.
\newblock \emph{Journal of Biomedical Informatics}, 70:\penalty0 35–51, June 2017.
\newblock ISSN 1532-0464.
\newblock \doi{10.1016/j.jbi.2017.04.009}.
\newblock URL \url{http://dx.doi.org/10.1016/j.jbi.2017.04.009}.

\bibitem[Claxton et~al.(2019)Claxton, Lutsey, MacLehose, Chen, Lewis, and Alonso]{Claxton2019}
J’Neka~S. Claxton, Pamela~L. Lutsey, Richard~F. MacLehose, Lin~Y. Chen, Tené~T. Lewis, and Alvaro Alonso.
\newblock Geographic disparities in the incidence of stroke among patients with atrial fibrillation in the united states.
\newblock \emph{Journal of Stroke and Cerebrovascular Diseases}, 28\penalty0 (4):\penalty0 890–899, April 2019.
\newblock ISSN 1052-3057.
\newblock \doi{10.1016/j.jstrokecerebrovasdis.2018.12.005}.
\newblock URL \url{http://dx.doi.org/10.1016/j.jstrokecerebrovasdis.2018.12.005}.

\bibitem[Dong et~al.(2021)Dong, Takeoka, Xiao, and Oyamada]{Dong2021}
Yuyang Dong, Kunihiro Takeoka, Chuan Xiao, and Masafumi Oyamada.
\newblock Efficient joinable table discovery in data lakes: A high-dimensional similarity-based approach.
\newblock In \emph{2021 IEEE 37th International Conference on Data Engineering (ICDE)}. IEEE, April 2021.
\newblock \doi{10.1109/icde51399.2021.00046}.
\newblock URL \url{http://dx.doi.org/10.1109/ICDE51399.2021.00046}.
\newblock \url{https://arxiv.org/abs/2010.13273}.

\bibitem[Dong et~al.(2023)Dong, Xiao, Nozawa, Enomoto, and Oyamada]{Dong2023}
Yuyang Dong, Chuan Xiao, Takuma Nozawa, Masafumi Enomoto, and Masafumi Oyamada.
\newblock Deepjoin: Joinable table discovery with pre-trained language models.
\newblock \emph{Proceedings of the VLDB Endowment}, 16\penalty0 (10):\penalty0 2458–2470, June 2023.
\newblock ISSN 2150-8097.
\newblock \doi{10.14778/3603581.3603587}.
\newblock URL \url{http://dx.doi.org/10.14778/3603581.3603587}.

\bibitem[Fan et~al.(2023)Fan, Wang, Li, Zhang, and Miller]{Fan2023}
Grace Fan, Jin Wang, Yuliang Li, Dan Zhang, and Ren\'{e}e~J. Miller.
\newblock Semantics-aware dataset discovery from data lakes with contextualized column-based representation learning.
\newblock \emph{Proc. VLDB Endow.}, 16\penalty0 (7):\penalty0 1726–1739, mar 2023.
\newblock ISSN 2150-8097.
\newblock \doi{10.14778/3587136.3587146}.
\newblock URL \url{https://doi.org/10.14778/3587136.3587146}.

\bibitem[Franklin et~al.(2020)Franklin, Chari, Foreman, Seneviratne, Gruen, McCusker, Das, and McGuinness]{Franklin2020-dg}
Jay D~S Franklin, Shruthi Chari, Morgan~A Foreman, Oshani Seneviratne, Daniel~M Gruen, James~P McCusker, Amar~K Das, and Deborah~L McGuinness.
\newblock Knowledge extraction of cohort characteristics in research publications.
\newblock \emph{AMIA Annu. Symp. Proc.}, 2020:\penalty0 462--471, 2020.

\bibitem[Gehrmann et~al.(2023)Gehrmann, Herczog, Decker, and Beyan]{Gehrmann2023}
Julia Gehrmann, Edit Herczog, Stefan Decker, and Oya Beyan.
\newblock What prevents us from reusing medical real-world data in research.
\newblock \emph{Scientific Data}, 10\penalty0 (1), July 2023.
\newblock ISSN 2052-4463.
\newblock \doi{10.1038/s41597-023-02361-2}.
\newblock URL \url{http://dx.doi.org/10.1038/s41597-023-02361-2}.

\bibitem[Glicksberg et~al.(2018)Glicksberg, Miotto, Johnson, Shameer, Li, Chen, and Dudley]{Glicksberg2018-vf}
Benjamin~S Glicksberg, Riccardo Miotto, Kipp~W Johnson, Khader Shameer, Li~Li, Rong Chen, and Joel~T Dudley.
\newblock Automated disease cohort selection using word embeddings from electronic health records.
\newblock \emph{Pac. Symp. Biocomput.}, 23:\penalty0 145--156, 2018.

\bibitem[Gokhale et~al.(2020)Gokhale, Chandan, Toulis, Gkoutos, Tino, and Nirantharakumar]{Gokhale2020}
Krishna~Margadhamane Gokhale, Joht~Singh Chandan, Konstantinos Toulis, Georgios Gkoutos, Peter Tino, and Krishnarajah Nirantharakumar.
\newblock Data extraction for epidemiological research (dexter): a novel tool for automated clinical epidemiology studies.
\newblock \emph{European Journal of Epidemiology}, 36\penalty0 (2):\penalty0 165–178, August 2020.
\newblock ISSN 1573-7284.
\newblock \doi{10.1007/s10654-020-00677-6}.
\newblock URL \url{http://dx.doi.org/10.1007/s10654-020-00677-6}.

\bibitem[Gérardin et~al.(2022)Gérardin, Mageau, Mékinian, Tannier, and Carrat]{Grardin2022}
Christel Gérardin, Arthur Mageau, Arsène Mékinian, Xavier Tannier, and Fabrice Carrat.
\newblock Construction of cohorts of similar patients from automatic extraction of medical concepts: Phenotype extraction study.
\newblock \emph{JMIR Medical Informatics}, 10\penalty0 (12):\penalty0 e42379, December 2022.
\newblock ISSN 2291-9694.
\newblock \doi{10.2196/42379}.
\newblock URL \url{http://dx.doi.org/10.2196/42379}.

\bibitem[Hai et~al.(2023)Hai, Koutras, Quix, and Jarke]{Hai2023}
Rihan Hai, Christos Koutras, Christoph Quix, and Matthias Jarke.
\newblock Data lakes: A survey of functions and systems.
\newblock \emph{IEEE Transactions on Knowledge and Data Engineering}, 35\penalty0 (12):\penalty0 12571–12590, December 2023.
\newblock ISSN 2326-3865.
\newblock \doi{10.1109/tkde.2023.3270101}.
\newblock URL \url{http://dx.doi.org/10.1109/TKDE.2023.3270101}.

\bibitem[Hripcsak et~al.(2015)Hripcsak, Duke, Shah, Reich, Huser, Schuemie, Suchard, Park, Wong, Rijnbeek, van~der Lei, Pratt, Nor{\'e}n, Li, Stang, Madigan, and Ryan]{Hripcsak2015-gf}
George Hripcsak, Jon~D Duke, Nigam~H Shah, Christian~G Reich, Vojtech Huser, Martijn~J Schuemie, Marc~A Suchard, Rae~Woong Park, Ian Chi~Kei Wong, Peter~R Rijnbeek, Johan van~der Lei, Nicole Pratt, G~Niklas Nor{\'e}n, Yu-Chuan Li, Paul~E Stang, David Madigan, and Patrick~B Ryan.
\newblock Observational health data sciences and informatics ({OHDSI)}: Opportunities for observational researchers.
\newblock \emph{Stud. Health Technol. Inform.}, 216:\penalty0 574--578, 2015.

\bibitem[Johnson et~al.(2016)Johnson, Pollard, Shen, Lehman, Feng, Ghassemi, Moody, Szolovits, Anthony~Celi, and Mark]{Johnson2016}
Alistair~E.W. Johnson, Tom~J. Pollard, Lu~Shen, Li-wei~H. Lehman, Mengling Feng, Mohammad Ghassemi, Benjamin Moody, Peter Szolovits, Leo Anthony~Celi, and Roger~G. Mark.
\newblock Mimic-iii, a freely accessible critical care database.
\newblock \emph{Scientific Data}, 3\penalty0 (1), May 2016.
\newblock ISSN 2052-4463.
\newblock \doi{10.1038/sdata.2016.35}.
\newblock URL \url{http://dx.doi.org/10.1038/sdata.2016.35}.

\bibitem[Kandula et~al.(2011)Kandula, Zeng-Treitler, Chen, Salomon, and Bray]{Kandula2011}
Sasikiran Kandula, Qing Zeng-Treitler, Lingji Chen, William~L. Salomon, and Bruce~E. Bray.
\newblock A bootstrapping algorithm to improve cohort identification using structured data.
\newblock \emph{Journal of Biomedical Informatics}, 44:\penalty0 S63–S68, December 2011.
\newblock ISSN 1532-0464.
\newblock \doi{10.1016/j.jbi.2011.10.013}.
\newblock URL \url{http://dx.doi.org/10.1016/j.jbi.2011.10.013}.

\bibitem[Kang et~al.(2017)Kang, Zhang, Tang, Hruby, Rusanov, Elhadad, and Weng]{Kang2017}
Tian Kang, Shaodian Zhang, Youlan Tang, Gregory~W Hruby, Alexander Rusanov, Noémie Elhadad, and Chunhua Weng.
\newblock Eliie: An open-source information extraction system for clinical trial eligibility criteria.
\newblock \emph{Journal of the American Medical Informatics Association}, 24\penalty0 (6):\penalty0 1062–1071, April 2017.
\newblock ISSN 1527-974X.
\newblock \doi{10.1093/jamia/ocx019}.
\newblock URL \url{http://dx.doi.org/10.1093/jamia/ocx019}.

\bibitem[Khatiwada et~al.(2022)Khatiwada, Shraga, Gatterbauer, and Miller]{Khatiwada2022}
Aamod Khatiwada, Roee Shraga, Wolfgang Gatterbauer, and Renée~J. Miller.
\newblock Integrating data lake tables.
\newblock \emph{Proceedings of the VLDB Endowment}, 16\penalty0 (4):\penalty0 932–945, December 2022.
\newblock ISSN 2150-8097.
\newblock \doi{10.14778/3574245.3574274}.
\newblock URL \url{http://dx.doi.org/10.14778/3574245.3574274}.

\bibitem[Kirby et~al.(2016)Kirby, Speltz, Rasmussen, Basford, Gottesman, Peissig, Pacheco, Tromp, Pathak, Carrell, Ellis, Lingren, Thompson, Savova, Haines, Roden, Harris, and Denny]{Kirby2016}
Jacqueline~C Kirby, Peter Speltz, Luke~V Rasmussen, Melissa Basford, Omri Gottesman, Peggy~L Peissig, Jennifer~A Pacheco, Gerard Tromp, Jyotishman Pathak, David~S Carrell, Stephen~B Ellis, Todd Lingren, Will~K Thompson, Guergana Savova, Jonathan Haines, Dan~M Roden, Paul~A Harris, and Joshua~C Denny.
\newblock Phekb: a catalog and workflow for creating electronic phenotype algorithms for transportability.
\newblock \emph{Journal of the American Medical Informatics Association}, 23\penalty0 (6):\penalty0 1046–1052, March 2016.
\newblock ISSN 1067-5027.
\newblock \doi{10.1093/jamia/ocv202}.
\newblock URL \url{http://dx.doi.org/10.1093/jamia/ocv202}.

\bibitem[Knight et~al.(2020)Knight, Spencer-Bonilla, Maahs, Blum, Valencia, Zuma, Prahalad, Sarraju, Rodriguez, and Scheinker]{Knight2020}
Gabriel~M Knight, Gabriela Spencer-Bonilla, David~M Maahs, Manuel~R Blum, Areli Valencia, Bongeka~Z Zuma, Priya Prahalad, Ashish Sarraju, Fatima Rodriguez, and David Scheinker.
\newblock Multimethod, multidataset analysis reveals paradoxical relationships between sociodemographic factors, hispanic ethnicity and diabetes.
\newblock \emph{BMJ Open Diabetes Research \& Care}, 8\penalty0 (2):\penalty0 e001725, November 2020.
\newblock ISSN 2052-4897.
\newblock \doi{10.1136/bmjdrc-2020-001725}.
\newblock URL \url{http://dx.doi.org/10.1136/bmjdrc-2020-001725}.

\bibitem[Kreuzthaler et~al.(2015)Kreuzthaler, Schulz, and Berghold]{Kreuzthaler2015}
Markus Kreuzthaler, Stefan Schulz, and Andrea Berghold.
\newblock Secondary use of electronic health records for building cohort studies through top-down information extraction.
\newblock \emph{Journal of Biomedical Informatics}, 53:\penalty0 188–195, February 2015.
\newblock ISSN 1532-0464.
\newblock \doi{10.1016/j.jbi.2014.10.010}.
\newblock URL \url{http://dx.doi.org/10.1016/j.jbi.2014.10.010}.

\bibitem[Li et~al.(2020)Li, Li, Suhara, Doan, and Tan]{Li2020}
Yuliang Li, Jinfeng Li, Yoshihiko Suhara, AnHai Doan, and Wang-Chiew Tan.
\newblock Deep entity matching with pre-trained language models.
\newblock \emph{Proceedings of the VLDB Endowment}, 14\penalty0 (1):\penalty0 50–60, September 2020.
\newblock ISSN 2150-8097.
\newblock \doi{10.14778/3421424.3421431}.
\newblock URL \url{http://dx.doi.org/10.14778/3421424.3421431}.

\bibitem[Liao et~al.(2015)Liao, Cai, Savova, Murphy, Karlson, Ananthakrishnan, Gainer, Shaw, Xia, Szolovits, Churchill, and Kohane]{Liao2015}
K.~P. Liao, T.~Cai, G.~K. Savova, S.~N. Murphy, E.~W. Karlson, A.~N. Ananthakrishnan, V.~S. Gainer, S.~Y. Shaw, Z.~Xia, P.~Szolovits, S.~Churchill, and I.~Kohane.
\newblock Development of phenotype algorithms using electronic medical records and incorporating natural language processing.
\newblock \emph{BMJ}, 350\penalty0 (apr24 11):\penalty0 h1885–h1885, April 2015.
\newblock ISSN 1756-1833.
\newblock \doi{10.1136/bmj.h1885}.
\newblock URL \url{http://dx.doi.org/10.1136/bmj.h1885}.

\bibitem[Miotto et~al.(2016)Miotto, Li, Kidd, and Dudley]{Miotto2016}
Riccardo Miotto, Li~Li, Brian~A. Kidd, and Joel~T. Dudley.
\newblock Deep patient: An unsupervised representation to predict the future of patients from the electronic health records.
\newblock \emph{Scientific Reports}, 6\penalty0 (1), May 2016.
\newblock ISSN 2045-2322.
\newblock \doi{10.1038/srep26094}.
\newblock URL \url{http://dx.doi.org/10.1038/srep26094}.

\bibitem[Mugambi et~al.(2023)Mugambi, Sherman, Fiterau, and Carreiro]{Mugambi2023}
Purity Mugambi, Michael Sherman, Madalina Fiterau, and Stephanie Carreiro.
\newblock 148: Disparities in ami treatment in the icu: Insights from multiple datasets.
\newblock \emph{Critical Care Medicine}, 52\penalty0 (1):\penalty0 S49–S49, December 2023.
\newblock ISSN 0090-3493.
\newblock \doi{10.1097/01.ccm.0000998784.35648.05}.
\newblock URL \url{http://dx.doi.org/10.1097/01.ccm.0000998784.35648.05}.

\bibitem[Narayan et~al.(2022)Narayan, Chami, Orr, and Ré]{Narayan2022}
Avanika Narayan, Ines Chami, Laurel Orr, and Christopher Ré.
\newblock Can foundation models wrangle your data?
\newblock \emph{Proceedings of the VLDB Endowment}, 16\penalty0 (4):\penalty0 738–746, December 2022.
\newblock ISSN 2150-8097.
\newblock \doi{10.14778/3574245.3574258}.
\newblock URL \url{http://dx.doi.org/10.14778/3574245.3574258}.

\bibitem[Pivovarov et~al.(2015)Pivovarov, Perotte, Grave, Angiolillo, Wiggins, and Elhadad]{Pivovarov2015}
Rimma Pivovarov, Adler~J. Perotte, Edouard Grave, John Angiolillo, Chris~H. Wiggins, and Noémie Elhadad.
\newblock Learning probabilistic phenotypes from heterogeneous ehr data.
\newblock \emph{Journal of Biomedical Informatics}, 58:\penalty0 156–165, December 2015.
\newblock ISSN 1532-0464.
\newblock \doi{10.1016/j.jbi.2015.10.001}.
\newblock URL \url{http://dx.doi.org/10.1016/j.jbi.2015.10.001}.

\bibitem[Pollard et~al.(2018)Pollard, Johnson, Raffa, Celi, Mark, and Badawi]{Pollard2018}
Tom~J. Pollard, Alistair E.~W. Johnson, Jesse~D. Raffa, Leo~A. Celi, Roger~G. Mark, and Omar Badawi.
\newblock The eicu collaborative research database, a freely available multi-center database for critical care research.
\newblock \emph{Scientific Data}, 5\penalty0 (1), September 2018.
\newblock ISSN 2052-4463.
\newblock \doi{10.1038/sdata.2018.178}.
\newblock URL \url{http://dx.doi.org/10.1038/sdata.2018.178}.

\bibitem[Reimers and Gurevych(2019)]{reimers-2019-sentence-bert}
Nils Reimers and Iryna Gurevych.
\newblock Sentence-bert: Sentence embeddings using siamese bert-networks.
\newblock In \emph{Proceedings of the 2019 Conference on Empirical Methods in Natural Language Processing}. Association for Computational Linguistics, 11 2019.
\newblock URL \url{http://arxiv.org/abs/1908.10084}.

\bibitem[Reisman(2017)]{Reisman2017-ea}
Miriam Reisman.
\newblock {EHRs}: The challenge of making electronic data usable and interoperable.
\newblock \emph{P T}, 42\penalty0 (9):\penalty0 572--575, September 2017.

\bibitem[Sarwar et~al.(2022)Sarwar, Seifollahi, Chan, Zhang, Aksakalli, Hudson, Verspoor, and Cavedon]{Sarwar2022}
Tabinda Sarwar, Sattar Seifollahi, Jeffrey Chan, Xiuzhen Zhang, Vural Aksakalli, Irene Hudson, Karin Verspoor, and Lawrence Cavedon.
\newblock The secondary use of electronic health records for data mining: Data characteristics and challenges.
\newblock \emph{ACM Computing Surveys}, 55\penalty0 (2):\penalty0 1–40, January 2022.
\newblock ISSN 1557-7341.
\newblock \doi{10.1145/3490234}.
\newblock URL \url{http://dx.doi.org/10.1145/3490234}.

\bibitem[Segura-Bedmar and Raez(2019)]{SeguraBedmar2019}
Isabel Segura-Bedmar and Pablo Raez.
\newblock Cohort selection for clinical trials using deep learning models.
\newblock \emph{Journal of the American Medical Informatics Association}, 26\penalty0 (11):\penalty0 1181–1188, September 2019.
\newblock ISSN 1527-974X.
\newblock \doi{10.1093/jamia/ocz139}.
\newblock URL \url{http://dx.doi.org/10.1093/jamia/ocz139}.

\bibitem[Suhara et~al.(2022)Suhara, Li, Li, Zhang, Demiralp, Chen, and Tan]{Suhara2022}
Yoshihiko Suhara, Jinfeng Li, Yuliang Li, Dan Zhang, undefinedağatay Demiralp, Chen Chen, and Wang-Chiew Tan.
\newblock Annotating columns with pre-trained language models.
\newblock In \emph{Proceedings of the 2022 International Conference on Management of Data}, SIGMOD/PODS ’22. ACM, June 2022.
\newblock \doi{10.1145/3514221.3517906}.
\newblock URL \url{http://dx.doi.org/10.1145/3514221.3517906}.

\bibitem[Toseef et~al.(2022)Toseef, Li, and Wong]{Toseef2022}
Muhammad Toseef, Xiangtao Li, and Ka-Chun Wong.
\newblock Reducing healthcare disparities using multiple multiethnic data distributions with fine-tuning of transfer learning.
\newblock \emph{Briefings in Bioinformatics}, 23\penalty0 (3), March 2022.
\newblock ISSN 1477-4054.
\newblock \doi{10.1093/bib/bbac078}.
\newblock URL \url{http://dx.doi.org/10.1093/bib/bbac078}.

\bibitem[WDC(2015)]{WDC2015}
WDC.
\newblock Web data commons (wdc) data corpus, 2015.
\newblock URL \url{https://webdatacommons.org/webtables/2015/downloadInstructions.html}.

\bibitem[Wieder and Nolte(2022)]{Wieder2022-kz}
Philipp Wieder and Hendrik Nolte.
\newblock Toward data lakes as central building blocks for data management and analysis.
\newblock \emph{Front. Big Data}, 5:\penalty0 945720, August 2022.

\bibitem[Xiong et~al.(2019)Xiong, Shi, Chen, Jiang, Tang, Wang, Chen, and Yan]{Xiong2019}
Ying Xiong, Xue Shi, Shuai Chen, Dehuan Jiang, Buzhou Tang, Xiaolong Wang, Qingcai Chen, and Jun Yan.
\newblock Cohort selection for clinical trials using hierarchical neural network.
\newblock \emph{Journal of the American Medical Informatics Association}, 26\penalty0 (11):\penalty0 1203–1208, July 2019.
\newblock ISSN 1527-974X.
\newblock \doi{10.1093/jamia/ocz099}.
\newblock URL \url{http://dx.doi.org/10.1093/jamia/ocz099}.

\bibitem[Yu et~al.(2015)Yu, Liao, Shaw, Gainer, Churchill, Szolovits, Murphy, Kohane, and Cai]{Yu2015}
Sheng Yu, Katherine~P Liao, Stanley~Y Shaw, Vivian~S Gainer, Susanne~E Churchill, Peter Szolovits, Shawn~N Murphy, Isaac~S. Kohane, and Tianxi Cai.
\newblock Toward high-throughput phenotyping: unbiased automated feature extraction and selection from knowledge sources.
\newblock \emph{Journal of the American Medical Informatics Association}, 22\penalty0 (5):\penalty0 993–1000, April 2015.
\newblock ISSN 1067-5027.
\newblock \doi{10.1093/jamia/ocv034}.
\newblock URL \url{http://dx.doi.org/10.1093/jamia/ocv034}.

\bibitem[Yu et~al.(2020)Yu, Chen, Yu, Li, Yang, Jiang, and Jiang]{yu-etal-2020-dataset}
Xiaojing Yu, Tianlong Chen, Zhengjie Yu, Huiyu Li, Yang Yang, Xiaoqian Jiang, and Anxiao Jiang.
\newblock Dataset and enhanced model for eligibility criteria-to-{SQL} semantic parsing.
\newblock In Nicoletta Calzolari, Fr{\'e}d{\'e}ric B{\'e}chet, Philippe Blache, Khalid Choukri, Christopher Cieri, Thierry Declerck, Sara Goggi, Hitoshi Isahara, Bente Maegaard, Joseph Mariani, H{\'e}l{\`e}ne Mazo, Asuncion Moreno, Jan Odijk, and Stelios Piperidis, editors, \emph{Proceedings of the Twelfth Language Resources and Evaluation Conference}, pages 5829--5837, Marseille, France, May 2020. European Language Resources Association.
\newblock ISBN 979-10-95546-34-4.
\newblock URL \url{https://aclanthology.org/2020.lrec-1.714}.

\bibitem[Yuan et~al.(2019)Yuan, Ryan, Ta, Guo, Li, Hardin, Makadia, Jin, Shang, Kang, and Weng]{Yuan2019}
Chi Yuan, Patrick~B Ryan, Casey Ta, Yixuan Guo, Ziran Li, Jill Hardin, Rupa Makadia, Peng Jin, Ning Shang, Tian Kang, and Chunhua Weng.
\newblock Criteria2query: a natural language interface to clinical databases for cohort definition.
\newblock \emph{Journal of the American Medical Informatics Association}, 26\penalty0 (4):\penalty0 294–305, February 2019.
\newblock ISSN 1527-974X.
\newblock \doi{10.1093/jamia/ocy178}.
\newblock URL \url{http://dx.doi.org/10.1093/jamia/ocy178}.

\bibitem[Zhang and Ives(2020)]{Zhang2020}
Yi~Zhang and Zachary~G. Ives.
\newblock Finding related tables in data lakes for interactive data science.
\newblock In \emph{Proceedings of the 2020 ACM SIGMOD International Conference on Management of Data}, SIGMOD/PODS ’20. ACM, May 2020.
\newblock \doi{10.1145/3318464.3389726}.
\newblock URL \url{http://dx.doi.org/10.1145/3318464.3389726}.

\bibitem[Zhu et~al.(2016)Zhu, Nargesian, Pu, and Miller]{Zhu2016}
Erkang Zhu, Fatemeh Nargesian, Ken~Q. Pu, and Ren\'{e}e~J. Miller.
\newblock Lsh ensemble: internet-scale domain search.
\newblock \emph{Proc. VLDB Endow.}, 9\penalty0 (12):\penalty0 1185–1196, aug 2016.
\newblock ISSN 2150-8097.
\newblock \doi{10.14778/2994509.2994534}.
\newblock URL \url{https://doi.org/10.14778/2994509.2994534}.

\bibitem[Zhu et~al.(2019)Zhu, Deng, Nargesian, and Miller]{Zhu2019}
Erkang Zhu, Dong Deng, Fatemeh Nargesian, and Renée~J. Miller.
\newblock Josie: Overlap set similarity search for finding joinable tables in data lakes.
\newblock In \emph{Proceedings of the 2019 International Conference on Management of Data}, SIGMOD/PODS ’19. ACM, June 2019.
\newblock \doi{10.1145/3299869.3300065}.
\newblock URL \url{http://dx.doi.org/10.1145/3299869.3300065}.

\end{thebibliography}

\section{First Appendix: Descriptions of columns of interest used in the main experiment}
\label{appendix-1}

\begin{table}[h!]
\caption{Description of columns of interest from MIMIC-III}
\centering
\resizebox{0.5\textwidth}{!}{\begin{tabular}{|p{1.5cm}|p{1.9cm}|p{1.45cm}|p{3cm}|}
\hline
\textbf{\makecell{Column\\name}} & \textbf{\makecell{Table\\Name(s)}} &  \textbf{\makecell{Data\\type}} & \textbf{Description}\\
\hline
subject\_id & admissions, patients & integer & unique patient id\\
\hline
hadm\_id & admissions, patients, prescriptions & long integer & unique admission id\\
\hline
admittime & admissions & timestamp & hospital admission time \\
\hline
\makecell{admission\\location} & admissions & text & hospital admission location \\
\hline
\makecell{discharge\\location} & admissions & text & hospital discharge location \\
\hline
insurance & admissions & text & type of insurance \\
\hline
ethnicity & admissions & text & race and ethnicity \\
\hline
diagnosis & admissions & text & initial diagnosis \\
\hline
seq\_num & diagnoses\_icd & integer & sequence of diagnosis priority \\
\hline
icd9\_code & diagnoses\_icd & text & diagnosis icd9 code \\
\hline
value & chartevents & text & contains several patient event data including past medical history \\
\hline
gender & patients & text & sex of the patient \\
\hline
dob & patients & timestamp & date of birth \\
\hline
drug & prescriptions & text & drug name \\
\hline
route & prescriptions & text & drug administration route \\
\hline
dose\_val\_rx & prescriptions & mixed& drug dosage \\
\hline
\end{tabular}}
\label{appendix-table:1}
\end{table}

\section{Second Appendix: Additional Usecases}
\subsection{Usecase 1}
\textbf{Research question}: Are the drugs and labs most commonly ordered for patients with AMI similar across datasets?

\textbf{Inclusion criteria}: Top 20 most ordered drugs and procedures for patients whose primary diagnosis is AMI.

Breaking down the inclusion criteria into queries, we identify the following tables and columns: \textit{prescriptions-table}, \textit{drug-name-column}, \textit{labs-table}, \textit{lab-name-column}, \textit{diagnosis-table}, \textit{diagnosis-column}, \textit{diagnosis-priority-column}, \textit{procedures-table}, and \textit{procedure-name-column}. In MIMIC-III these tables and columns map to: prescriptions, drug, d\_labitems, label, diagnosis\_icd, diagnosis, icd9\_code, seq\_num, d\_icd\_procedures, and, long\_title respectively. 
\subsection{Usecase 2}
\textbf{Research question}: Is the number of patient clinical notes associated with disease severity? 

\textbf{Inclusion criteria}: Adult patients that experienced shock during their hospitalization and/or died in hospital, and their associated notes.

The following tables and columns are identified as being important to answering this question. \textit{Diagnosis-table}, \textit{diagnosis-icd9-code-column}, \textit{admissions-table}, \textit{discharge-location-column}, \textit{noteevents-table}, and \textit{note-text-column}. In MIMIC-III these tables and columns map to: diagnosis\_icd, icd9\_code, admissions, discharge\_location, noteevents, and text respectively.

Table \ref{table:extra-usecases} summarizes the columns of interest in these additional usecases and their matches in eICU.

\begin{table}[htbp]
\caption{Reference columns and their matches in eICU.}
\centering
\resizebox{0.49\textwidth}{!}{\begin{tabular}{|l|l|l|}
\hline
\textbf{\makecell{MIMIC\\column name}} & \textbf{\makecell{eICU\\column name}} &  \textbf{\makecell{eICU\\table name}}\\
\hline
\hline
drug & drugname & medication \\
\hline
label & labname & lab \\
\hline
icd9\_code & icd9code & diagnosis\\
\hline
diagnosis & diagnosisstring & diagnosis\\
\hline
seq\_num & diagnosispriority $^{\mathrm{a}}$ & diagnosis\\
\hline
long\_title & treatmentstring$^{\mathrm{b}}$ & treatment\\
\hline
discharge\_location & hospitaldischargelocation & patient\\
\hline
text$^{\mathrm{c}}$ & \makecell{[nursingchartvalue, cellattributevalue, \\ notetext, physicalexamtext]} & \makecell{[nursingcharting, nurseassessment, \\note, physicalexam]}\\
\hline
\multicolumn{3}{p{430pt}}{$^{\mathrm{a}}$ Diagnosis priority is recorded through sequence numbers (integers) in MIMIC-III, however, this same information in eICU is recorded as short text (string). They contain the necessary information to identify diagnosis priority but are very different structurally.} \\
\multicolumn{3}{p{430pt}}{$^{\mathrm{b}}$ eICU does not have a dedicated procedures table, like MIMIC-III does. Some times, procedure information is included \emph{treatmentstring} within the \emph{treatment} table. Therefore, this is not a true match, but at best a partial match.}\\
\multicolumn{3}{p{430pt}}{$^{\mathrm{c}}$ There are several tables in eICU that could contain patient care notes. These are typically much smaller and less detailed than the \emph{text} column in the \emph{noteevents} table in MIMIC-III.}\\
\end{tabular}}
\label{table:extra-usecases}
\end{table}

\begin{figure}[htbp]
\includegraphics[width=\columnwidth]{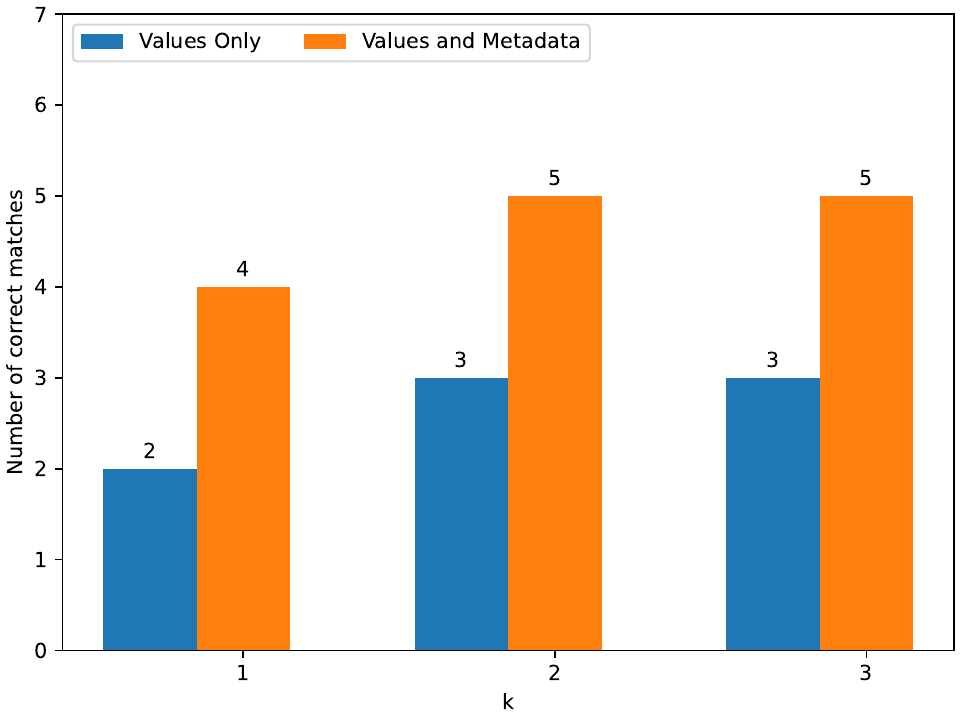}
\caption{Accuracy of top $3$ matches as the number of additional columns increases. $N$(total number of columns evaluated) $=7$}
\label{fig:extra-usecases-results}
\end{figure}

We matched the reference columns against the \emph{entire} eICU database. Top-three matches by values-only and when metadata are included are summarized in Table \ref{table:extra-usecases-results} and in Fig \ref{fig:extra-usecases-results}.

\begin{table}[htbp]
\caption{Top-three matches by column-values only and when metadata are included.}
\centering
\resizebox{0.49\textwidth}{!}{\begin{tabular}{|l|l|l|}
\hline
\textbf{\makecell{MIMIC\\column name}} & \textbf{\makecell{Matches by \\value-only}} & \textbf{\makecell{Matches by\\value and metadata}}\\
\hline
\hline
drug & drugname, allergyname, antibiotic & drugname, allergyname, dosage \\
\hline
label & \makecell{labresultrevisedoffset, drugname, \\ allergyname} & \makecell{labresultrevisedoffset, labname, \\ labmeasurenamesystem} \\
\hline
icd9\_code & \makecell{cpleolid, nursecareentryoffset, \\ intakeoutputentryoffset} & \makecell{cpleolid, nursecareentryoffset, \\intakeoutputentryoffset}\\
\hline
diagnosis & \makecell{specialty, diagnosisstring, \\ physicianspeciality} & \makecell{diagnosisstring, admitdiagnosis, \\ usertype} \\
\hline
seq\_num & infusionrate, bun, respiratoryrate & infusionrate, bun, respiratoryrate \\
\hline
long\_title & dischargeweight, culturesite, specialty & \makecell{physicianspeciality, specialty, \\ diagnosisstring} \\
\hline
discharge\_location & \makecell{hospitaldischargelocation, cplitemvalue, \\ cellattribute} & hospitaldischargelocation, usertype, drugname \\
\hline
text & \makecell{dischargeweight, cplgoalvalue, \\ unittype} & \makecell{physicalexamtext, dischargeweight, \\ pasthistorynotetype} \\
\hline
\multicolumn{3}{p{380pt}}{$^{\mathrm{a}}$ These columns names have ``hospital" preceding them in the database.} \\
\end{tabular}}
\label{table:extra-usecases-results}
\end{table}

\section{Third Appendix: Errors, suggested improvements, and computation time}
\subsection{Errors and suggested improvements}
In this section, we present examples of instances when the proposed algorithm fails.

\textbf{Case 1: Columns with similar values are matched, despite not being contextually related.}
When matching columns based on column values alone, the proposed method fails to correctly match subject\_id, hadm\_id, icd9\_code, value, gender, and dose\_val\_rx within the top 3 returned matches (Table \ref{table:3}). In MIMIC-III, subject\_id, hadm\_id, icd9\_code, are integer values which can map to other columns (e.g., highly ranked columns customlabid, nursingchartid, noteid, patientunitstayid) with integer values. Value-wise, these are correct matches, i.e., integer values are matched to other integer values, but context-wise, they are incorrect. These errors are corrected when metadata such as, tables in which columns co-occur, are provided. This helps the algorithm find \textit{columns whose values are integers \textbf{and} that co-occur in specific tables}. Including other complex metadata such as whether column is a primary/foreign key, can further help in getting these matches correct.

\textbf{Case 2: Data inconsistency - mixed column value types, and different formatting.} 
The algorithm fails to correctly match value, and gender columns when using column values alone (Table \ref{table:3}) because of inconsistent data types and values. In MIMIC-III, gender is recorded as 'F', 'M', while in eICU it is recorded as 'Female' and 'Male'. We are able to correctly match columns when column names are included in the matching. On the other hand, values in the column \emph{value} are of mixed datatypes. This column represents information about diverse sets of data that are recorded in clinicial charts. For instance, vitals information such as temperature and blood pressure, and past medical history would be recorded in this same column. When using values alone, the algorithm finds that it matches columns with varied data types some containing text data and others numerical data. 

\textbf{Case 3: Unavailable matches}
For several columns (indicated in grey in Table \ref{table:3}), there are no actual matches in eICU. This is expected because based on database designs, some information may be aggregated in a single column in one database while separated into several databases in another, e.g., using admittime and dob to compute patient age in MIMIC-III, while the 'age' column directly exists in eICU. The algorithm will find matches for these columns based on data types and metadata, despite not being ``true" matches for that column. This is a tricky case for the researchers because they do not know whether the algorithm failed or whether those columns have no matches in the other database. The database schemas can be used to disambiguate this, and future work will look into how this schema information can be integrated in the algorithm to offer further support to the researcher.

\subsection{Computation time}
As stated in section 3.2.1, the proposed algorithm is a runtime Linear to the number of columns in the database. The most expensive cost the algorithm pays is generating embeddings for all values in each column (line 2 in Algorithm \ref{algorithm-1}). This cost however is not prohibitive when using a smaller model such as SBERT. For instance, using 2 GPUs, we generated embeddings for the largest column (column name: value, table: Chartevents, dataset: MIMIC-III) which has 316,158,414 values in 10hours. For columns with fewer unique values, e.g., ethnicity and gender, it takes 9.5 and 5.6 seconds respectively, to read and encode unique values in all the batches. Furthermore, the cost of encoding values is paid once, henceforth, the embeddings are used to compute distances and find matches for various columns extracted from different criteria.

\section{Fourth Appendix: Strawman version - naive column-name-based matching.}
\begin{table}[h!]
\caption{Top results on matches based on embeddings of column names. The correct matches have been bolded.}
    \centering
    \resizebox{\textwidth}{!}{\begin{tabular}{|l|l|}
    \hline
    \textbf{\makecell{MIMIC\\column name}} & \textbf{\makecell{Top matches from eICU with decreasing similarity score}}\\
    \hline
    subject\_id & usertype(0.719), admissiondxid(0.7164), allergyid(0.7159), writtenineicu(0.7158), apacheapsvarid(0.7153)\\
    \hline
    hadm\_id & \makecell[l]{admissiondxid(0.9027), admitdxenteredoffset(0.9027), admitdxname(0.9021), allergyid(0.9019),\\ admitdxpath(0.9004)}\\
    \hline
    discharge\_location & \makecell[l]{dischargelocation(0.9931), region(0.9926), albumin(0.9926), urine(0.9925), acutephysiologyscore(0.9925), \\amilocation(0.9925), \textbf{hospitaldichargelocation(0.9925)}, dischargeweight(0.9925),
    wbc(0.9925), \\cplcareprovderid(0.9925), ejectfx(0.9925), unitdischargelocation(0.9925)}\\
    \hline
    admission\_location	& \makecell[l]{admissiondxid(0.993), admitdxpath(0.991), region(0.9904), admitdxname(0.9904), admitdxtext(0.9904), \\ albumin(0.9904), visitnumber(0.9904)}\\
    \hline
    admittime & \makecell[l]{admitdxtext(0.9585), admitdxname(0.9582), admitdxenteredoffset(0.957), admissiondxid(0.9554),\\ admitdxpath(0.9551)}\\
    \hline
    diagnosis & \makecell[l]{admitdiagnosis(0.9985), acutephysiologyscore(0.9983), diagnosisid(0.9983), predictedhospitalmortality(0.9983), \\\textbf{diagnosisstring(0.9983})}\\
    \hline
    gender	& \textbf{gender(1.0)}, ethnicity(1.0), st3(1.0), st2(1.0), pvr(1.0)\\
    \hline
    ethnicity &	\textbf{ethnicity(0.9974)}, gender(0.9974), region(0.9973), preopmi(0.9973), age(0.9973)\\
    \hline
    route & \textbf{routeadmin(1.0)}, icp(1.0), pasthistorypath(1.0), sao2(1.0), st3(1.0)\\
    \hline
    icd9\_code & \makecell[l]{\textbf{icd9code(0.9999)}, icp(0.9999), cplinfectdiseaseoffset(0.9999), patienthealthsystemstayid(0.9999),\\ hospitaldischargetime24(0.9999)}\\
    \hline
    value	& - \\
    \hline
    drug & drugrate(1.0), drugamount(1.0), drugvadmixture(1.0), drugorderoffset(1.0), drugstartoffset(1.0)\\
    \hline
    dose\_val\_rx	& \textbf{dosage(1.0)}, treatmentid(1.0), pvri(1.0), treatmentoffset(1.0), treatmentstring(1.0) \\
    \hline
    \end{tabular}}
    \label{appendix-table:2}
\end{table}

\end{document}